\documentclass{article}

\usepackage{microtype}
\usepackage{graphicx}
\usepackage{subfigure}
\usepackage{booktabs} % for professional tables
\usepackage{listings}

\usepackage{hyperref}
\usepackage{xcolor}
\hypersetup{colorlinks,linkcolor={red}}

\usepackage[accepted]{icml2024}

\usepackage{amsmath}
\usepackage{amssymb}
\usepackage{mathtools}
\usepackage{amsthm}
\usepackage{algorithm}
\usepackage{amsthm}
\usepackage{svg}
\usepackage{svg}
\usepackage{multirow}
\usepackage{bbding}
\usepackage{listings}
\usepackage{xcolor}
\usepackage{verbatim}
\usepackage{enumitem}

\usepackage[capitalize,noabbrev]{cleveref}

\theoremstyle{plain}

\theoremstyle{definition}

\theoremstyle{remark}

\usepackage[textsize=tiny]{todonotes}

\newcommand{\eg}{\emph{e.g.}}

\usepackage{lipsum}
\newcommand\blfootnote[1]{%
  \begingroup
  \renewcommand\thefootnote{}\footnote{#1}%
  \addtocounter{footnote}{-1}%
  \endgroup
}

\def \robotlogos {\raisebox{-0.1\height}{\includegraphics[height=0.9\baselineskip]{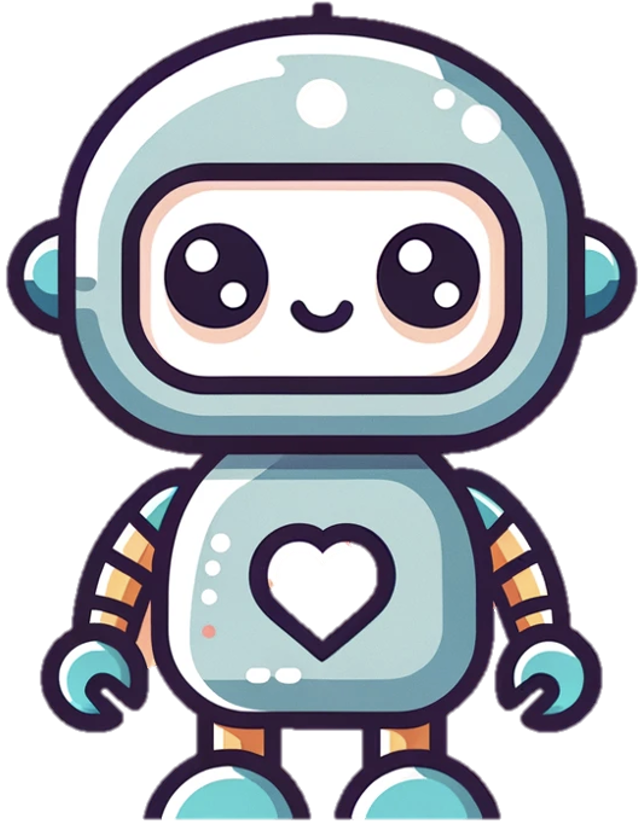}}}

\usepackage{xspace}
\newcommand{\alias}{RoboCodeX\xspace}
\begin{document}

\twocolumn[
\icmltitle{\robotlogos{} RoboCodeX: Multimodal Code Generation for Robotic Behavior Synthesis}

\centering{\small
	\textbf{Yao Mu$^{12\dag}$,
Junting Chen$^{23}$,
Qinglong Zhang$^{2}$,
Shoufa Chen$^{1}$,
Qiaojun Yu$^{4}$,
Chongjian Ge$^{1}$,
Runjian Chen$^{1}$,\\
Zhixuan Liang$^{1}$,
Mengkang Hu$^{1}$,
Chaofan Tao$^{1}$,
Peize Sun$^{1}$,
Haibao Yu$^{1}$,
Chao Yang$^{2}$,
Wenqi Shao$^{2}$,\\
Wenhai Wang$^{5}$,
Jifeng Dai$^{2,6}$,
Yu Qiao$^{2}$,
Mingyu Ding$^{7*}$,
Ping Luo$^{1,2*}$} \\
\vspace{10pt}
	\small $^1$The University of Hong Kong, \quad $^2$OpenGVLab, Shanghai AI Laboratory \quad $^3$ETH Zurich  \quad $^4$Shanghai Jiao Tong University  \\
 $^5$The Chinese University of Hong Kong \quad $^6$Tsinghua University \quad $^7$UC Berkely \\
	{\url{https://sites.google.com/view/robocodexplus}} \\
 
}
% \vspace{10pt}

\textbf{
\centerline{\emph{Talk is cheap. Show me the code.}}
\rightline{--- Linus Torvalds}}

\vskip 0.3in
]

\blfootnote{
$\dag$ Project lead. 
* Corresponding authors.}
\begin{abstract}

Robotic behavior synthesis, the problem of understanding multimodal inputs and generating precise physical control for robots, is an important part of Embodied AI.
Despite successes in applying multimodal large language models for high-level understanding, it remains challenging to translate these conceptual understandings into detailed robotic actions while achieving generalization across various scenarios.
In this paper, we propose a tree-structured multimodal code generation framework for generalized robotic behavior synthesis, termed \alias.
\alias decomposes high-level human instructions into multiple object-centric manipulation units consisting of physical preferences such as affordance and safety constraints, and applies code generation to introduce generalization ability across various robotics platforms.
% enhance the capability to map visual understanding into physical 
To further enhance the capability to map conceptual and perceptual understanding into control commands, a specialized multimodal reasoning dataset is collected for pre-training and an iterative self-updating methodology is introduced for supervised fine-tuning.
Extensive experiments demonstrate that \alias achieves state-of-the-art performance in both simulators and real robots on four different kinds of manipulation tasks and 
% competitive performance on 
one navigation task. 
% More demos and information can be found in our 
% \href{https://sites.google.com/view/robocodex}{homepage}.

\begin{figure}[t]
  \centering
   \includegraphics[width=0.99\linewidth]{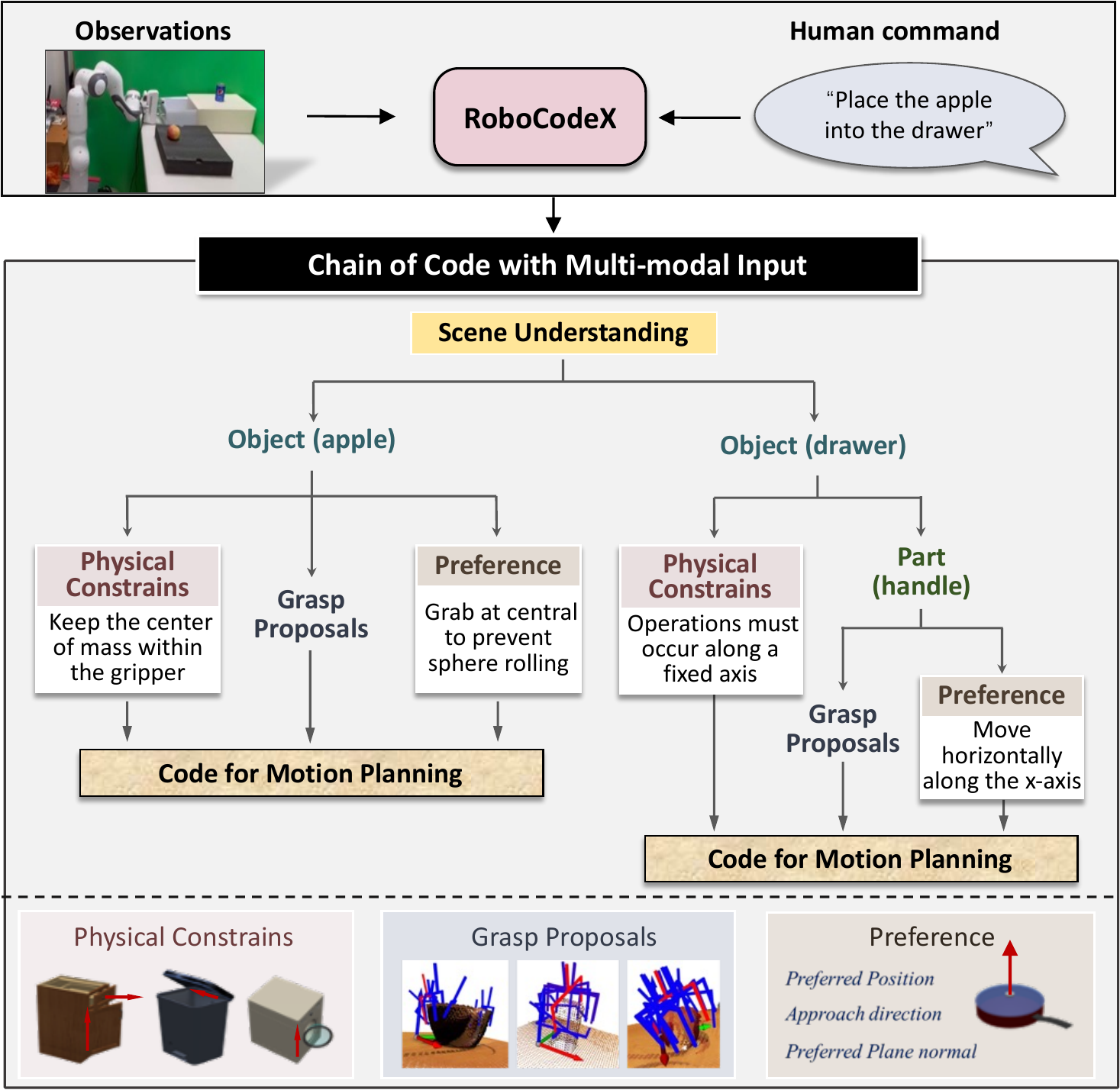}
   \vspace{-8pt}
   \caption{An illustration of our \alias \robotlogos{}, a large vision language model with tree-of-thought reasoning capabilities for robotic code generation.
   It decomposes high-level human instructions into multiple object-focused manipulation subtasks, further expanding them by predicting physical constraints, preferential rankings, and target position proposals.
   \alias acts as an interface to translate high-level semantic understanding of the observed environment and instructions into tailored robotic behaviors. 
   %
   % The framework utilizes a tree-of-thought approach to decompose high-level instructions into multiple object-focused manipulation subtasks. 
   % %
   % Each subtask functions as a node in the tree decomposition.
   % %
   % The nodes are further expanded to predict proposals for target object positions, physical properties, preferential rankings of proposals, and feasible robot trajectories to complete each subtask. 
   % \my{TBD, three bottom figures} 
   }
   \vspace{-10pt}
   \label{fig:teaser}
\end{figure}

\end{abstract}
    
\section{Introduction}
\label{sec:intro}

Embodied AI equips intelligent agents, such as robots, with the capacity for perception, reasoning, and interaction within the 3D physical world and with humans. 
Yet, a central challenge lies in the generalizability of robotic manipulation frameworks.
On one hand, robot policies may struggle to generalize to objects with diverse properties and mechanisms, \eg, objects with different physical constraints and grasping preferences in Figure.~\ref{fig:teaser}.
On the other hand, adapting a robot learning framework to different robotic platforms encounters difficulties. 
These challenges stem from the gap between the high-level scene understanding and the low-level manipulation and control policies.
How to create a general framework with reasoning capabilities that can map the high-level semantic understanding to generic robotics behaviors, thus becomes a primary concern.

Previous methods~\cite{saycan,huang2023grounded,rana2023sayplan} leverage the reasoning capabilities of Large Language Models (LLMs) to propose step-by-step natural language plans for robot execution. 
However, lacking environmental information (\eg, status and shape of objects from vision) as contextual grounding, it is difficult to leverage LLMs for decision-making within a given real-world context. Consequently, they rely heavily on low-level skill models~\cite{brohan2022rt, rt2} to convert visual observations and language commands into low-level actions. 
Nonetheless, such skill models still struggle to generalize to novel scenarios or robotic platforms, which requires precise tuning of action sequences tailored to the robot's morphology and actuators. 

Specifically, robotic skills and low-level behaviors for different tasks encompass these precise action sequences that fulfill proper contact points, grasping poses, physical constraints, waypoints, and movement paths.
Though previous methods~\cite{gao2023physically,mirjalili2023lan,agrawal2023physical, Mo_2019_CVPR,xu2022universal,eisner2022flowbot3d,zhong20233d,li2023manipllm} have explored estimating manipulation preferences, such as grasp regions of an object~\cite{mirjalili2023lan} and contact points~\cite{li2023manipllm}, their focuses are mainly on specific tasks and objects. Coordinating and generalizing such modules to diverse common tasks, such as multiple object interactions in complex environments, remains a significant challenge.

Recently, the rapid advancement of Multimodal Large Language Models (MLLMs)~\cite{li2022blip,alayrac2022flamingo,liu2023improved,liu2023visual}, such as GPT-4V~\cite{openai2023gpt4}, have demonstrated strong aptitude in scene comprehension, understanding of objects' characteristics, and generalization capabilities across diverse scenes and objects. This suggests that leveraging their cognitive strengths by establishing connections to physical world interactions could enable more efficient intelligent system design. Additionally, while MLLMs excel at high-level scene understanding, robotic systems specialize in precise trajectory planning and execution. This implies that code could act as a symbolic bridge between conceptual knowledge and low-level behaviors for robots. Such connections would enable the integration of MLLMs' advanced cognitive processing with the nuanced requirements of robotic control and planning, potentially leading to more foundational and adaptable cross-platform robotic systems.

Inspired by the above observations, 
in this paper, we propose a large vision language model for robotic code generation, named \alias, to serve as an interface between MLLMs and robotic control systems, translating high-level semantics and physical preference into corresponding low-level motions tailored to a robot's mechanics. 
Critically, representing these plans and preferences in code enables sharing and transfer across morphologically distinct robots.

Specifically, \alias comprehends diverse sensory inputs like vision, depth data, and human instructions, and employs a tree-of-thought structure that decomposes high-level human instructions into several object-centric manipulation units defined through textual code annotations. 
Each unit acts as a node, further broken down into several key components: 
\textbf{1)} Target pose proposals generation through detailed 3D spatial analysis of objects within their environments. 
\textbf{2)} Object physical characteristic prediction with visual inspiration and 3D geometry information such as translation or rotation axis constraints on articulated objects, ensuring the generation of realistic and feasible motion plans. 
\textbf{3)} Preference prediction with the multimodal understanding of environmental and task objectives, selecting optimal gripping positions and approach directions. 
\textbf{4)} Trajectory generation leverages the planning algorithm to produce safe, optimized, and compliant motion plans to respect collision and physical constraints. 
With the seamlessly integrated key capabilities, \alias performs complex manipulation tasks flexibly in cluttered physical environments based on high-level commands.

Training an MLLM with such reasoning capacity for physical robotic preferences and robotic behavior in codes remains a challenge, which is often underrepresented in standard pretraining datasets primarily focusing on high-level visual and language alignment.
Meanwhile, research on reasoning by fusing information from multiple sources is seldom explored.
To address this gap, we construct a multimodal reasoning dataset for pretraining by generating simulated environments with diverse tasks and corresponding executable codes for each task. 
For the supervised fine-tuning~(SFT) stage, we generate high-quality data and code with high success rate via an iterative self-updating framework.
The key contributions of this work are as follows:

\begin{itemize}[leftmargin=10pt, itemsep=-2pt]
\vspace{-12pt}
\item We introduce \alias, a large vision language model with tree-of-thought reasoning capabilities for robotic code generation. It predicts physical constraints, preferential rankings, and target position proposals while serving as an interface between high-level conceptual knowledge and low-level robotic behaviors.
\item We present a specialized multimodal reasoning dataset and an iterative self-updating methodology for supervised fine-tuning to enhance \alias's capacity for translating semantics and physical preferences into robot-specific motions.
\item Extensive experiments demonstrate state-of-the-art performance of \alias in both simulated and real robot systems across four different kinds of manipulation tasks, \eg, 17\% success rate improvement over GPT-4V, and competitive performance on embodied navigation tasks.

\end{itemize}

\begin{figure*}[t]
  \centering
  % \fbox{\rule{0pt}{2in} \rule{0.9\linewidth}{0pt}}
   \includegraphics[width=1.0\linewidth]{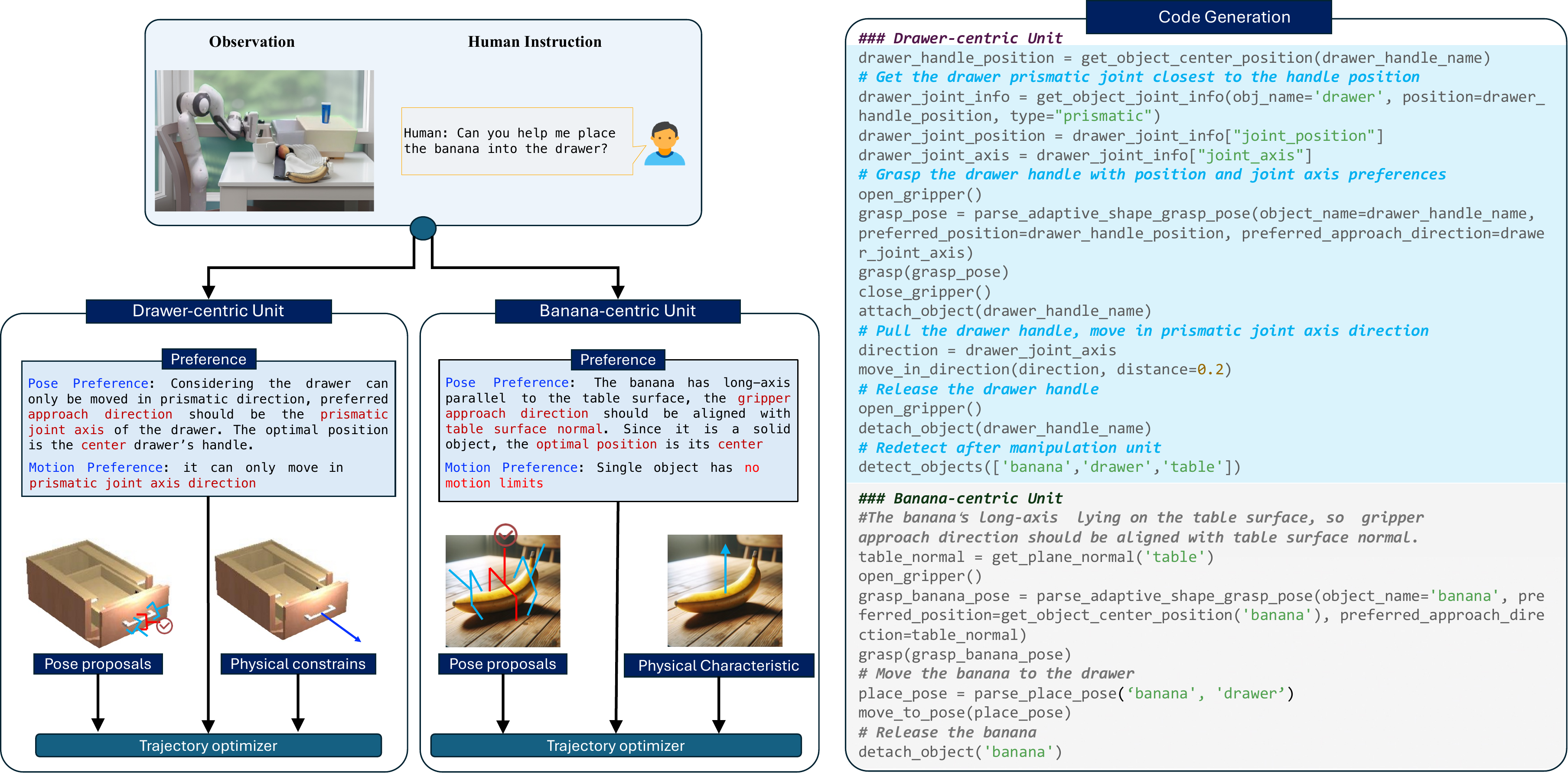}
   % \vspace{-8pt}
   \caption{Example of robotic behavior synthesis with RoboCodeX. To accomplish the task ``Place the banana into the drawer," we first decompose the whole task into a Drawer-centric Unit and a Banana-centric Unit. In the Drawer-centric Unit, the robot is programmed to understand that it must align its gripper with the prismatic joint axis of the drawer, which is the optimal position for movement, considering the drawer's physical limits and trajectory optimization. Conversely, the Banana-centric Unit requires the robot to align its gripper with the table surface normal and close to its center to pick up a banana. The accompanying code generation segment translates these multimodal considerations into executable instructions. For the drawer, the code includes determining the handle's position, executing the grip and pull actions in line with the drawer’s joint axis, and then releasing the handle. For the banana, the code sequences involve aligning the gripper, grasping the banana, moving it to the drawer, and detaching it at the destination. }
   \label{fig:model}
   \vspace{-4pt}
\end{figure*}

\section{Related Works}
\noindent \textbf{Code Generation for Robotic Control.}
Recent advances have explored utilizing natural language models for robotic behavior synthesis. ProgPrompt~\cite{progprompt2023} introduced in the context of virtual environments, leverages natural language models to generate action sequences in VirtualHome~\cite{puig2018virtualhome}, enabling intuitive and manageable robot behavior synthesis tailored for domestic settings. Building upon this, Code-as-Policies~\cite{codeaspolicies2022} investigates composing full policy code for manipulation and navigation tasks by emphasizing few-shot prompting and hierarchical code generation. This work incorporates modular structures for reasoning and control into the generated programs. Further developments by~\citep{vemprala2023chatgpt} focus these techniques on real robotic systems. By mapping high-level functions to diverse atomic tasks, they rapidly adapt capabilities across robotic form factors while generating tailored executable code. However, these methods rely solely on linguistic inputs, lacking multimodal grounding to map visual observations to behaviors. Different from previous works, this paper achieves visually-grounded code synthesis to generate behaviors specialized to perceived object characteristics. 
\noindent \textbf{Embodied AI with Large Foundation Models.}
Leveraging large pre-trained models shows promise for creating capable embodied agents. Many works focus on using language models for planning and reasoning in embodied agents~\cite{huang2022language,saycan,chen2023open,progprompt2023,huang2023visual,raman2022planning,song2023llm,zhou2023generalizable,hu2023tree,liu2023llm+,vemprala2023chatgpt,ding2023task,palme,yuan2023plan4mc,lu2023multimodal,wang2023voyager,radford2019language,padalkar2023open}. To enable language models to perceive physical environments, common approaches include providing textual descriptions of scenes~\cite{huang2022inner,zeng2022socratic,progprompt2023} or access to perception APIs~\cite{codeaspolicies2022}. Vision can also be incorporated by decoding with visual context~\cite{huang2023grounded} or using multi-modal language models that directly take visual input~\cite{palme,OpenAI2023GPT4TR,embodiedgpt,yang2023octopus}. Several researchers further adapt visual models to the embodied perspective under egocentric conditions \cite{nair2022r3m,huang2018predicting} and build datasets for bridging the asynchronous egocentric and third-person view of human activities in the real world \cite{grauman2022ego4d,huang2023egobridge}.
However, perception alone is not enough for embodied agents - they must also know how to act. This is often achieved by providing pre-defined action primitives that the model can invoke. As shown in \cite{codeaspolicies2022}, LLMs can exhibit useful behavioral common sense for low-level control policies, in addition to high-level reasoning and planning capabilities. However, effectively connecting perception, reasoning, and grounded action primitives remains an open challenge. Unlike previous approaches that use separate pipelines for vision and language or only focus on high-level planning, our proposed method bridges the gap between the cognitive abilities of a large end-to-end multi-modal model and precise robotic planning through code generation. This approach allows the model to make multi-modal predictions, expanding each code node with information about target positions, physical properties, preferential rankings, and feasible trajectories.

\noindent \textbf{Multimodal Large Language Models.}
Recent advancements have seen the creation of multimodal large language models~(MLLMs)~\cite{zhang2023llama-adapter, zhang2023internlmxcomposer, zhang2023gpt4roi, wu2023nextgpt, sun2023emu, alayrac2022flamingo, zhu2023ghost, li2023videochat, lai2023lisa, yang2023gpt-4v, chen2022pali, li2023otter, zhang2023video-llama, li2023monkey, ye2023mplugdocowl}, which aim to enhance LLM with the capability to process and interpret visual information. Flamingo~\cite{alayrac2022flamingo} uses the visual and language inputs as prompts and shows remarkable few-shot performance for visual question answering.
Subsequently, LLaVA series~\cite{liu2023llava, lu2023empirical, liu2023improved} and MiniGPT-4~\cite{zhu2023minigpt4} have brought in visual instruction tuning, to improve the instruction-following ability of MLLMs.
Concurrently, models such as VisionLLM~\cite{wang2023visionllm}, KOSMOS-2~\cite{peng2023kosmos2}, and Qwen-VL~\cite{bai2023qwenvl}, along with other works~\cite{wang2023allseeing, chen2023shikra} have improved MLLMs with visual grounding capabilities, facilitating tasks such as region description and localization.
However, despite the considerable achievements of MLLMs, their multimodal reasoning capabilities in complex robotic behavior synthesizing remain under-explored.

\section{Methods}
\label{sec:meth}
We propose a multi-modal code generation framework that utilizes a tree-of-thought architecture to decompose instructions into code units expanded through multi-modal predictions. To enhance the reasoning ability of the MLLMs, we develop a specialized dataset and an iterative, specially designed fine-tuning methodology. We evaluate this framework with different preferences. Furthermore, we integrate a vision adapter into the vision-language model to facilitate multi-scale visual feature integration and bridge cognitive perceptions with robotic planning.
\subsection{Problem Setup}
We consider a long-term manipulation problem as a high-level free-form human instruction $\mathcal{L}_{\text{global}}$ (\eg, ``clean the table''). The observation contains the RGBD data $I$ from three different views of the depth camera~(left, right, top). Since it is hard to generate the whole trajectories directly for long-term high-level instructions, we use a tree-of-thought method via the visual reasoning of MLLMs to decompose the entire task into several object-centric individual units $\mathcal{U}_{task} \to (u_1, u_2, \dots, u_n)$.
The central problem investigated in this work is how to generate a motion trajectory set $\{\tau_i\}_{i=0}^{n}$. We represent $\tau_i$ as a sequence of dense end-effector waypoints to be executed by an Operational Space Controller~\cite{khatib1987unified}, where each waypoint consists of a desired 6-DoF end-effector pose, end-effector velocity, and gripper action.  Given the $i$-th sub-task described by ground-truth instruction $\ell_i^*$, we formulate an optimization problem defined as follows:
\begin{equation}
\vspace{-3pt}
\small
\begin{split}
\min _{\tau_i} \sum_{i=0}^N\left\{S_{\text {task }}\left(\tau_i, \ell_i^*\right)+S_{\text {control }}\left(\tau_i\right)\right\}  \quad \textit{s.t.} \quad \mathcal{C}(\mathbf{\tau}_i),\\
\end{split}
\vspace{-3pt}
\end{equation}
where $S_{\text{task}}$ scores the extent to which $\tau_i$ completes the instruction $\ell_i$, while $S_{\text{control}}$ specifies the control costs, \eg, the cost to encourage $\tau_i$ to minimize total control effort or total time. 
$\mathcal{C}(\mathbf{\tau}_i)$ denotes the dynamics and kinematics constraints. By solving this optimization for each sub-task $\ell_i$, we obtain a sequence of robot trajectories that collectively fulfill the overall task specified by the instruction $\mathcal{L}_{\text{global}}$.

\subsection{Multi-modal Tree-of-thought Code Generation}
We present a novel framework for synthesizing robotic behavior, integrating multi-perspective 3D perception, multi-modal chain-of-code reasoning, and motion planning for the execution of complex real-world tasks. The process begins by capturing observations \( I \), comprising 3 RGBD frames from 3 different views. This approach helps avoid occlusion issues caused by a single viewpoint and provides comprehensive 3D information. These are fused into a unified 3D spatial representation, \emph{i.e.}, a Truncated Signed Distance Field~(TSDF), encoding precise environmental structure.

Concurrently, the RGB images from three views are fed as inputs into the \alias model together with the natural language instructions. \alias employs a tree-of-thought architecture centered around semantic parsing, contextual grounding, and goal-oriented partitioning. Such visual reasoning directly connects visual features to an understanding of interaction preferences and physical constraints on objects. The overall task is decomposed into sequential object-centric units as demonstrated in Equation \eqref{decom} with the language description of the sub-task $L_i$ and the manipulation preference $pf_{i}$ which indicates the approach directions or preferred touching positions considering the control stability. For example, spherical objects should be grasped from the center for stability, while open containers like cups are better grasped at the edge for flexible control.
\begin{equation}
\small
\begin{gathered}
L_{\text {units }}, p f_{\text {units }}=f\left(I, L_{\text {global }}\right) \\
L_{\text {units }}=\left\{L_0, L_1, \ldots, L_N\right\} \\
p f_{\text {units }}=\left\{pf_{0}, pf_{1}, \ldots, pf_{N}\right\}.
\end{gathered}
\label{decom}
\end{equation}
For each object-centric unit, the grounded 2D positions derived from RGB streams are matched with 3D boxes from the point cloud, based on overlap and orientation consistency. This process enables the extraction of an accurate 3D point cloud, denoted as \( O_i \) for the task-relevant object. Subsequently, the object-centric trajectory generation problem, conditioned on the language instruction of the sub-task $L_{i}$ can be formulated as,
\begin{equation}
\small
\tau_i=g\left(O_i, L_i, p f_i\right),
\end{equation}
% \vspace{-3pt}
where \( \tau_i \) is the motion trajectory of the \( i \)-th sub-task generated by the function \( g(\cdot) \) with  the preference \( pf_i \) and language instruction $L_i$ for the \( i \)-th task. It contains a sequence of dense end-effector waypoints to be executed and discrete control instructions for gripper opening and closing.

Then the aggregated object-specific perceptual inferences, physical insights, and manipulation parameters are systematically compiled into structured executable action code. Each unit is considered as a parent node and is further expanded to \textit{\textbf{i) part-level affordance}} $A_i=h(O_i,L_i)$ prediction, which is the task-relevant part point cloud segmentation, such as the handle's area of the drawer; \textit{\textbf{ii) grasp pose proposals}} \( pc_{i} \) prediction, $p c_i=\left\{p c_{i 0}, \ldots, p c_{i j}, \ldots, p c_{i k}\right\}=\mathrm{cand}\left(A_i\right)$. Function $\text{cand}(\cdot)$ is a mapping $\mathbb{N}^3 \to \text{SE}(3)$ to generates SE$(3)$ poses. Specifically, we use pre-trained AnyGrasp~\cite{fang2023anygrasp} to generate the grasp pose proposals, which provide abundant candidates with scores (see details in Appendix~\ref{app:grasp}). We select the top 10 candidates as grasp pose proposals and use the preferences inferred by visual reasoning to select the optimal one, with consideration of the object's shape, prior knowledge, and the adjacent robot manipulation cells that will be performed. This leverages AnyGrasp's ability to produce diverse candidates while allowing task-specific selection via visual reasoning to pick high-quality ones; \textit{\textbf{iii) object physical property}} prediction \( \phi_i = \{\phi_{i0}, \ldots, \phi_{ij}, \ldots, \phi_{ik}\} = z(O_i, L_i) \) such as the joints information of the articulated objects. Specifically,  we use the GAMMA~\cite{yu2023gamma} model to predict the physical properties of articulated objects, which segments an articulated object point cloud into rigid parts and estimates articulation parameters. GAMMA extracts point-wise features using PointNet++~\cite{qi2017pointnet++} which are then processed for segmentation, part offset, and joint axis regression (see details in Appendix~\ref{app:joints}). We also encapsulate the plane detection functions in Open3D~\cite{zhou2018open3d} into an API for calling during code generation, taking the object's point cloud as input and outputting the detected planes as well as normal vector information; and \textit{\textbf{iv) trajectory planning}}:
By integrating motion planning algorithms and the Robot Operating System~(ROS) manipulation modules, the model finally outputs dynamically feasible robot trajectories with assurances of collision avoidance and singularity exclusion.
\begin{equation}
\small
\tau_i = \mathcal{Z}(pc_{i}, \phi_i, pf_{i}, L_i).
\end{equation}
Specifically, we employ zeroth-order optimization in trajectory planning by randomly sampling trajectories and evaluating them with $S_{\text{control}}$ and the constraint from the occupancy map (see example in~Appendix Figure~\ref{fig:occ_map}). 

\subsection{Dataset Preparation}
\textbf{Pre-training Dataset.}
Most multi-modal models currently focus on aligning different modalities, leaving a research gap in complex reasoning with information from multiple sources. To address this, we develop a diverse robotic multi-modal code generation dataset with the help of GPT-4.
To ensure the diversity of the dataset, we design a procedural data generation framework. We first randomly sample household scenes from the HM3D dataset~\cite{ramakrishnan2021habitat}, which provides various indoor scenes like bedrooms, living rooms, and kitchens as the base environments. We then insert additional objects into semantically appropriate locations in the scenes—for example, placing objects on top of tables and counters. The inserted objects are of two types: independent objects like balls, toys, and fruits; and container objects like bowls, plates, and cups that can hold other objects. The objects are sampled from Google Scan Dataset~\cite{downs2022google}, YCB Dataset~\cite{calli2015benchmarking}, OmniObject3D Dataset~\cite{wu2023omniobject3d}, and articulated object dataset AKB-48\cite{liu2022akb}. By randomly selecting object categories and quantities and populating the scenes accordingly, complex scene configurations are obtained. Built upon the resulting environments, we form natural language descriptions which are inputted into the  GPT-4 language model to produce free-form task descriptions suitable for the given scene configurations, specifying goals like object manipulation and rearrangement. 

Note that since we can obtain all the information from the simulator and render the observation from any view, we can use GPT-4 to generate data rather than GPT-4V, which is more expensive for large-scale data generation. Finally, for each generated task, programming codes that accomplish assigned tasks are generated by GPT-4 from the task descriptions and additional parameterized inputs. We sample 10 high-quality code samples for each task which evaluated by GPT-3.5  and filter out those with syntax errors, obtaining executable programs for the robot to carry out in simulation. 
Through this pipeline combining environment generation, task specification, and program synthesis with large language models, we build a diverse dataset for pre-training robotic vision language reasoning models to accomplish various daily household tasks. To avoid overfitting, we use both the general vision language pre-training dataset and the generated dataset together during the pre-training process. The general vision language pre-training dataset we use contains ShareGPT4V~\cite{chen2023sharegpt4v} dataset, SViT~\cite{zhao2023svit} dataset, and the LLaVA Visual Instruct 150K dataset~\cite{liu2023llava}. 

 \textbf{SFT Dataset.}
 The supervised fine-tuning (SFT) process~\cite{dong2023abilities} is meticulously designed to refine the model's proficiency in robotic code generation by training the model with a curated selection of high-quality code sequences that demonstrate a high success rate in evaluation. To create high-quality and diverse code data, we take the task types from the RT-1~\cite{brohan2022rt} and LIBERO~\cite{liu2023libero} datasets as a basis and randomly combine them with diverse objects to create a diverse set of tasks for each task type. For each type of task, we provide high-quality examples generated by humans, which have been verified to ensure successful completion in both simulator and real-world environments. We utilize GPT-4 to generate corresponding code for each task, taking high-quality examples and API explanations as input. For code with correct syntax but failed execution, we conduct an exhaustive search of optional settings, including grasping method (central lift or AnyGrasp), preferred contact position, preferred gripper direction, and trajectory generation method (fixed direction corresponding to objects' joints or only fixed final position). With the results of the search, we select the best option as the label and use GPT-4V to analyze why it is better and summarize it into coding annotations that serve as the label of the chain-of-thought. For the codes that maintain a zero success rate even after exploring optional settings, we manually revise the code and incorporate it into the example pool of human-labeled data. It's important to note that during the SFT process, we still require general vision language data to prevent overfitting. Specifically, we utilize the same subset of the ShareGPT4V dataset used in the ShareGPT4V-7B model's SFT process~\cite{chen2023sharegpt4v}.

\subsection{Vision Language Model Design}
Our model adopts the fundamental paradigm of BLIP2~\cite{{li2023blip}}, consisting of a vision transformer, Q-Former, and language model. 
As code generation for complex tasks tends to have considerable length and requires inputting long API documentation prompts, we leverage the Q-Former to bridge and compress the token quantity from the visual modality. Specifically, the vision transformer encodes visual elements into rich representational tokens. As full sequencing of these vision tokens alongside lengthy textual prompts may result in a high requirement of GPU memory, we adopt Q-Former to judiciously summarize the visual embeddings into more compact sequence tokens. These condensed representations capture the most task-relevant visual concepts while reducing sequence length to save GPU memory. The compressed Q-Former visual sequence tokens are then fed into the language model along with text tokens from code and documentation prompts. 
Furthermore, in order to obtain hierarchical features from the image, we design an efficient vision adapter that aggregates the features from different stages in the vision transformer. We first divide the hidden layers of the vision model into four parts proportionally. Then, the class token is extracted from the final layer of each part because it interacts with all tokens through self-attention within that layer. We concatenate these class tokens from the four hierarchical layers and enable interactions among them via the vision adapter. The vision adapter is a channel-wise attention network~\cite{yan2021channel}, which first reduces the channel dimension through a linear layer, then selects features using a SILU~\cite{paul2022sinlu} activation, and finally restores the original channel dimension using another linear layer (see pseudo-code in Appendix~\ref{app_model}). Finally, the aggregated feature tokens are concatenated with the other visual tokens. This combined set then serves as the input for the subsequent modules.

\begin{figure*}[t]
  \centering
  % \fbox{\rule{0pt}{2in} \rule{0.9\linewidth}{0pt}}
   \includegraphics[width=1.0\linewidth]{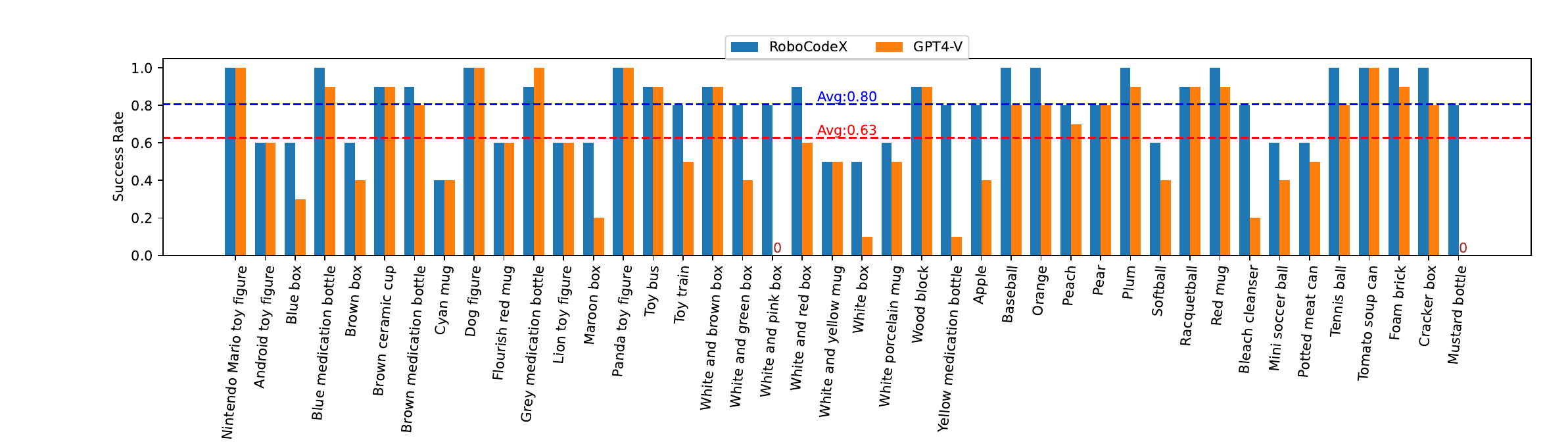}
   \vspace{-20pt}
   \caption{Performance Comparison on pick and place task with diverse objects.}
   \label{fig:result_pick_place}
   \vspace{-15pt}
\end{figure*}
\section{Experiments}

\subsection{Evaluation on Manipulation Task}
We evaluate our framework on a series of robotic manipulation tasks  with increasing complexity utilizing Gazebo~\cite{qian2014manipulation} for
the realistic rendering of the environment and the physical
interactions: 1) open-vocabulary picking-and-place task which involves 42 different object categories from the YCB and Google Scanned object datasets, 2) Opening and closing drawers from 5 typical cabinets taken from PartNet-Mobility dataset~\cite{xiang2020sapien}, 3) Opening and closing doors from the same set of 5 cabinets, 4) Placing objects in drawers, with the robot instructed to interact with the ordered drawer, and 5) Multi-stage tasks composed of sequential subtasks that need to be completed in order, such as putting all the fruits in the drawer(see detailed setup in Appendix \ref{app:setup}). 
The baselines consist of various multimodal models and large language models, including GPT-4V, GPT-3.5, and GPT-4, each equipped with open vocabulary object detection models. Among them, GPT-4V is a multimodal model taking images and language instructions as input, while GPT-4 and GPT-3.5 are language-only models, the inputs being text scene descriptions containing object names, attributes, and language instructions. For open vocabulary detection, we choose GLIP-Large~\cite{li2022grounded}, owing to its flexibility in handling diversified objects like ``tiny panda figure" in our tasks. We provide all the prompts and the explanation of the APIs for code generation in Appendix~\ref{app:api and prompt}.

\begin{figure}[t]
  \centering
\includegraphics[width=1.0\linewidth]{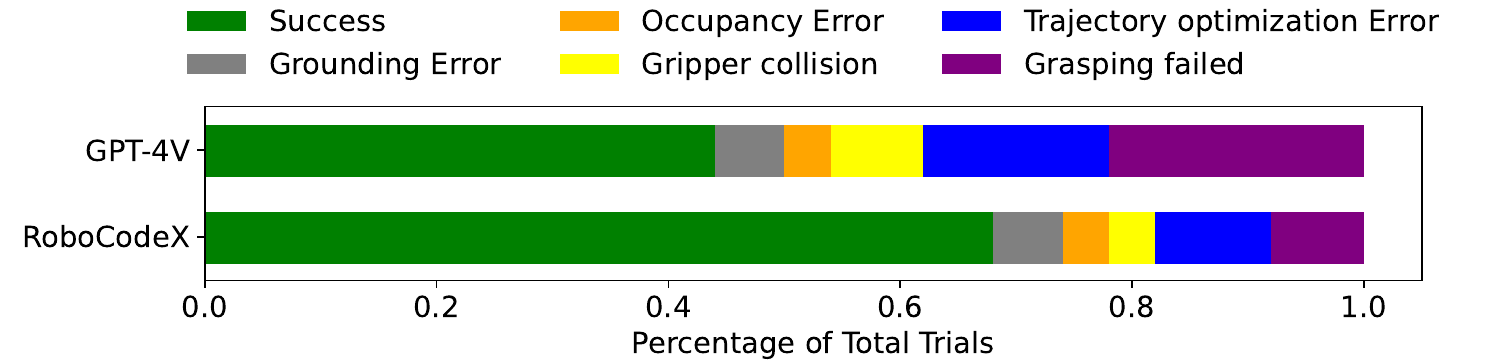}
\vspace{-15pt}
   \caption{Failure modes comparison between \alias and GPT-4V in long-term tasks.}
   \label{fig:errorbar}
   \vspace{-15pt}
\end{figure}

As shown in Figure~\ref{fig:result_pick_place}, \alias outperforms GPT-4V on pick-and-place tasks, primarily due to using more suitable priors for objects that GPT-4V struggled with. Compared to GPT-4, adding visual inputs enables GPT-4V to infer better grasp preferences. However, GPT-3.5 exhibited weaker reasoning abilities and often generated syntactically flawed text. As shown in Table~\ref{performace_sum}, \alias demonstrates stronger capabilities than GPT-4V on articulated object manipulation and long-term tasks. Its core competency lies in locating target objects and corresponding handles across multiple drawers and doors in the cabinet, and properly operating them under physical constraints. For example, determining whether the target cabinet door should be rotated clockwise or counterclockwise relies more heavily on visual understanding, which highlights \alias's advanced competencies in robotic manipulation beyond text-based interfaces.
Since long-term tasks involve many influencing factors, to gain a clear understanding of how much each factor plays into the whole long-term manipulation process, as shown in Figure~\ref{fig:errorbar}, we specifically conducted error analyses on \alias's and GPT-4V's long-term task performance. The results show both models can correctly understand and decompose these tasks. The primary errors stem from suboptimal grasp pose selection and trajectory planning failures, with the latter strongly correlated to grasp poses. Furthermore, \alias's multi-modal code generation training effectively reduces this error margin compared to GPT-4V.
\begin{table}[t]
\centering
% \small
\setlength{\tabcolsep}{3pt}
\caption{Performance comparison among different tasks. We report the average success rate of all the tasks over 50 trials.}
\resizebox{0.99\linewidth}{!}{%
\begin{tabular}{l|ccccc}
\toprule
                & \text{\footnotesize Pick \& Place} & \text{\footnotesize Drawers} & \text{\footnotesize Doors} & \text{\footnotesize Put obj. in drawer}  & \text{\footnotesize Multi-stage} \\               
\midrule
GPT-3.5  &0.45  & 0.44                 & 0.20               & 0.36   &  0.20      \\
GPT-4     &0.56  & 0.68                 & 0.46               & 0.48 &    0.44         \\
GPT-4V   &0.63  & 0.68                 & 0.52               & 0.55  & 0.50           \\
\alias      &\textbf{0.80}  & \textbf{0.84}                 & \textbf{0.74}               & \textbf{0.68}    &\textbf{0.64}          \\
\bottomrule
\end{tabular}
}
\label{performace_sum}
\end{table}

\begin{table}[t]
\centering
\vspace{-10pt}
\setlength{\tabcolsep}{10pt}
    \caption{Performance on Embodied Navigation Tasks. We report the success rate and Success weighted by Path Length (SPL) as the evaluation metrics\cite{anderson2018evaluation}.}
    \label{tab:results_navigation}

\resizebox{0.99\linewidth}{!}{%
\begin{tabular}{ccccc}
\toprule 
\multirow{2.5}{*}{ Method } & \multicolumn{2}{c}{ HM3D } & \multicolumn{2}{c}{ HSSD } \\
\cmidrule { 2 - 5 } & Success $\uparrow$ & SPL $\uparrow$ & Success $\uparrow$ & SPL $\uparrow$ \\
\midrule L3MVN(GPT2) & 35.2 & 16.5 & 38.4 & 19.4 \\
Pixel-Nav(GPT4) & 37.9 & 20.5 & - & - \\
ESC(GPT3.5) & 39.2 & 22.3 & - & - \\
\alias & $\mathbf{4 0 . 0}$ & $\mathbf{2 4 . 2}$ & $\mathbf{4 0 . 3}$ & $\mathbf{2 2 . 0}$ \\
\bottomrule
\end{tabular}
}
\vspace{-10pt}
\end{table}

\begin{table}[t]
    \centering
\footnotesize
    % \vspace{-15pt}
    % \setlength{\tabcolsep}{8pt}
    \caption{Performance on general multimodal reasoning.}
    \label{tab:results_reasoning}
    \resizebox{0.99\linewidth}{!}{%
    \setlength\tabcolsep{18pt} 
\begin{tabular}{l|cc}
\toprule Model & LLaVA-Bench & MM-Vet \\
\midrule BLIP-2 & 38.1 & 22.4 \\
InstructBLIP-7B & 60.9 & 26.2 \\
InstructBLIP-13B & 58.2 & 25.6 \\
LLaVA-1.5-7B & 63.4 & 30.5 \\
LLaVA-1.5-13B & 70.7 & \textbf{35.4}\\
\alias-13B & \textbf{71.5}& 31.0\\
\bottomrule
\end{tabular}
}
    \vspace{-15pt}
\end{table}

\begin{figure*}[t]
  \centering
   \includegraphics[width=1.0\linewidth]{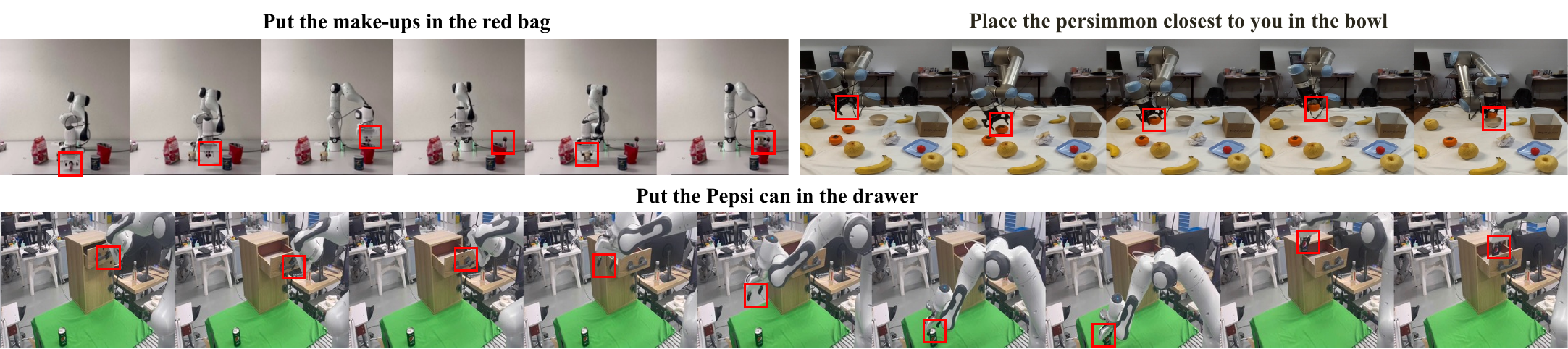}
   \vspace{-15pt}
   \caption{Generalization among different types of robots in real world without any fine-tuning. We evaluate \alias with Franka Emika Panda robot and UR5 robot in real world.}
   \vspace{-10pt}
   \label{fig:real-world}
\end{figure*}

\begin{figure}[t]
  \centering
\includegraphics[width=1.0\linewidth]{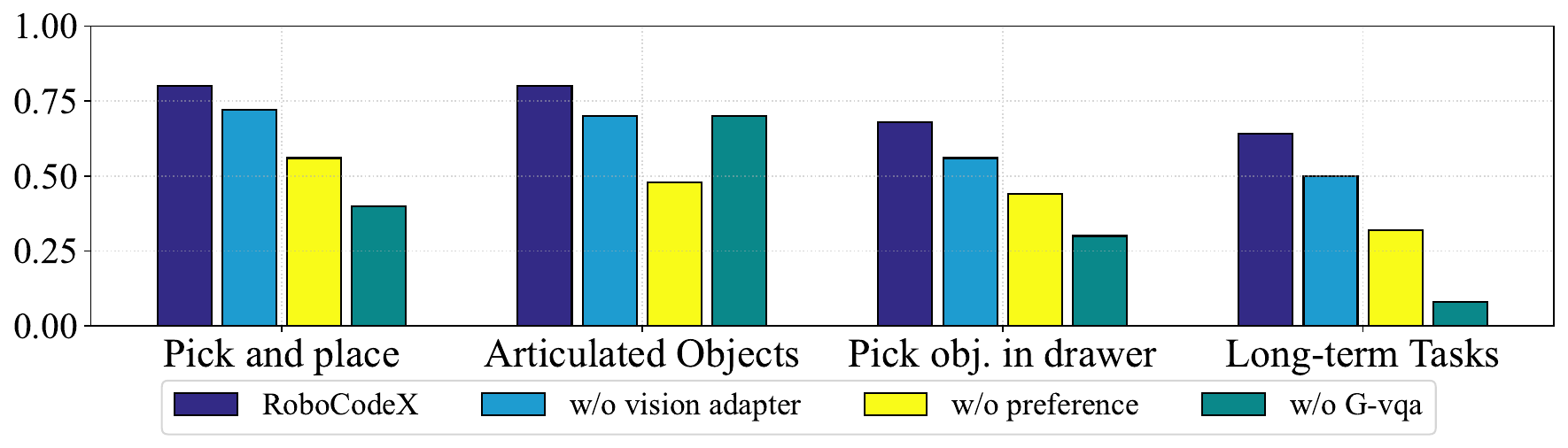}
\vspace{-15pt}
   \caption{Ablation on the utilization of preference, vision adapter, and whether to use general VQA data during fine-tuning. We report the average success rate over 4 kinds of tasks over 50 trials.}
   \label{fig:ablation}
   % \vspace{-10pt}
\end{figure}

\subsection{Evaluation on Embodied Navigation Task}

To test the performance of \alias on the visual language object-navigation task, we conduct experiments based on the HM3D~\cite{ramakrishnan2021habitat} and HSSD~\cite{khanna2023habitat} datasets.
For navigation tasks, \alias mainly utilizes the L3MVN~\cite{yu2023l3mvn} framework, where it utilizes a large language model to decide which frontier to select. We adopt this paradigm where \alias performs multimodal visual reasoning over the current observation and candidate frontiers and determines where to explore. We select the typical methods that use large foundation models to assist robots on visual language object navigation as key baselines, including L3MVN, Pixel-Nav~\cite{cai2023bridging} which analyzes panoramic images and utilizes GPT-4 to determine the optimal pixel for exploration, and ESC~\cite{zhou2023esc} which uses GPT-3.5~\cite{ouyang2022training} to determine the mid-term goal from the frontier points during exploration. As shown in Table~\ref{tab:results_navigation}, compared to these baselines, our method achieves better performance and is comparable to GPT-3.5, which demonstrates its general visual reasoning ability.

\subsection{Evaluation on General VQA}
To test \alias's general multimodal understanding and reasoning capabilities, we utilize two benchmarks:
1) LLaVA-Bench~\cite{liu2023llava}, a widely used benchmark for assessing multimodal conversation abilities. It scores model responses against reference responses using GPT-4.
2) MM-Vet~\cite{yu2023mm}, an evaluation benchmark focused on complicated multimodal tasks, designed with the insight that the ability to solve complex tasks is often achieved by integrating different core vision-language capabilities into a generalist model. As shown in Table~\ref{tab:results_reasoning}, \alias demonstrates comparable general visual reasoning to LLaVA-1.5-13B~\cite{liu2023improved}, the current state-of-the-art 13B multi-modal model, even without specialized fine-tuning on robotic tasks.

\subsection{Real World Experiments}

To validate the generalization ability of \alias to different real-world scenarios across different robot platforms, we test it on both the Franka Emika robot arm~\cite{haddadin2022franka} and the UR5 robot arm~\cite{kebria2016kinematic} which include multi-stage pick-and-place tasks, and put-object-in-drawer tasks, in a zero-shot way. As shown in Figure~\ref{fig:real-world}, the results show that our framework can adapt to different robots out-of-the-box simply by changing the robot's configuration file, demonstrating its strong generalization to new scenarios without any specific fine-tuning.

\subsection{Ablation Study}
We conduct ablation studies to evaluate the contributions of various components in the \alias's framework, which focus on examining the utility of the preference model, the vision adapter, and the incorporation of general visual question answering~(G-VQA) data during fine-tuning.

\textbf{Ablation on the preference.} 
We evaluate whether the inferred preference is necessary for robust manipulation, compared to directly using the highest-scoring grasp predicted by Anygrasp. As shown in Figure~\ref{fig:ablation}, the preference significantly outperforms the baseline across all the tasks, which highlights its efficacy in enhancing manipulation stability and aligning with subsequent planning.

\textbf{Ablation on vision adapter.}
To evaluate the effectiveness of the vision adapter, we conduct an experiment by removing it from the model. The results indicate that the vision adapter contributes to improving the success rate, primarily by aiding the model in better understanding the fine details of various objects. However, while it enhances performance, it is not the most critical component that determines the overall performance of the model.

\textbf{Ablation on general VQA data utilization.}
We explore the necessity of leveraging general VQA data during the SFT process. The results indicate that without general VQA (w/o G-VQA) data, the model exhibits significant overfitting. Its ability to follow instructions for different objects has deteriorated and it often answers questions improperly, therefore the average success rate has dropped significantly.

Overall, the ablation studies confirm the critical role of preference prediction and the vision adapter. These core components work together to enable precise object handling and adaptation to a variety of tasks. Furthermore, the incorporation of general VQA data during fine-tuning is essential to prevent overfitting issues.

\section{Conclusion}
\label{sec:con}
This work introduces \alias, a pioneering multi-modal code generation framework that bridges the gap between multi-modal large language models~(MLLMs) and robotic control systems, innovatively translating semantic understanding into tailored robotic behaviors.
Through extensive experimentation, \alias demonstrates state-of-the-art performance in complex robotic manipulation tasks, showcasing its robustness and adaptability in both simulated and real-world environments.
The integration of a multi-modal tree-of-thought approach, specialized dataset, and iterative fine-tuning methodology significantly enhances the model's capacity to interpret visual observation and human instructions into precise, robot-specific actions. Our findings suggest that the fusion of MLLMs' cognitive strengths with nuanced physical world interactions heralds a new era in embodied AI, where robots can efficiently adapt to and manipulate their environment with unprecedented sophistication. 
Future research could explore the expansion of \alias's capabilities to more diverse tasks, further unlocking the potential of multi-modal AI in robotics.

\section*{Acknowledgments}
We would like to express our profound gratitude to a distinguished group of individuals and teams whose exceptional contributions have been instrumental in advancing our project. We are particularly grateful to Professor Yufeng Yue and Dr. Guangyan Chen from the Beijing Institute of Technology for their expertise in real-world robot code deployment and their insightful feedback.
Special acknowledgment goes to Dr. Yifei Huang from the Shanghai AI Lab for his support with egocentric human-object interaction data and suggestions for pretraining egocentric vision language models. Our thanks also extend to Professor Cewu Lu, Dr. Wenhai Liu and Mr. Chenxi Wang from Shanghai Jiao Tong University for their support in integrating the Anygrasp model seamlessly into our system, which has been crucial in enhancing our project's functionality.
We are deeply appreciative of the support from the teams led by Professor Huazhe Xu and Professor Xueqian Wang at Tsinghua University. Their extensive assistance in robot hardware has been invaluable, allowing us to utilize cutting-edge technology in the development of our project. Furthermore, we are sincerely thankful to Dr. René Zurbrügg from the ETH AI Center for his assistance with the robot experimental platform. These collaborations have not only significantly enriched our project but also laid a foundation for future innovation. We extend our heartfelt thanks to everyone for their dedication, support, and invaluable contributions.

{
    \small
    \bibliographystyle{ref}
    \bibliography{main}
    
}

\clearpage
% $$ $$
% \newpage
\appendix

% \label{sec:con}

\section{Manipulation Simulation Setup}
\label{app:setup}
In this section, we delineate the methodology employed for constructing a sophisticated robotic simulation, integrating a Franka robotic arm within a dynamic environment composed of different cabinets sourced from the PartNet-Mobility Dataset and a table adorned with various objects.
\begin{figure}[htbp]
  \centering
  % \fbox{\rule{0pt}{2in} \rule{0.9\linewidth}{0pt}}
\includegraphics[width=0.85\linewidth]{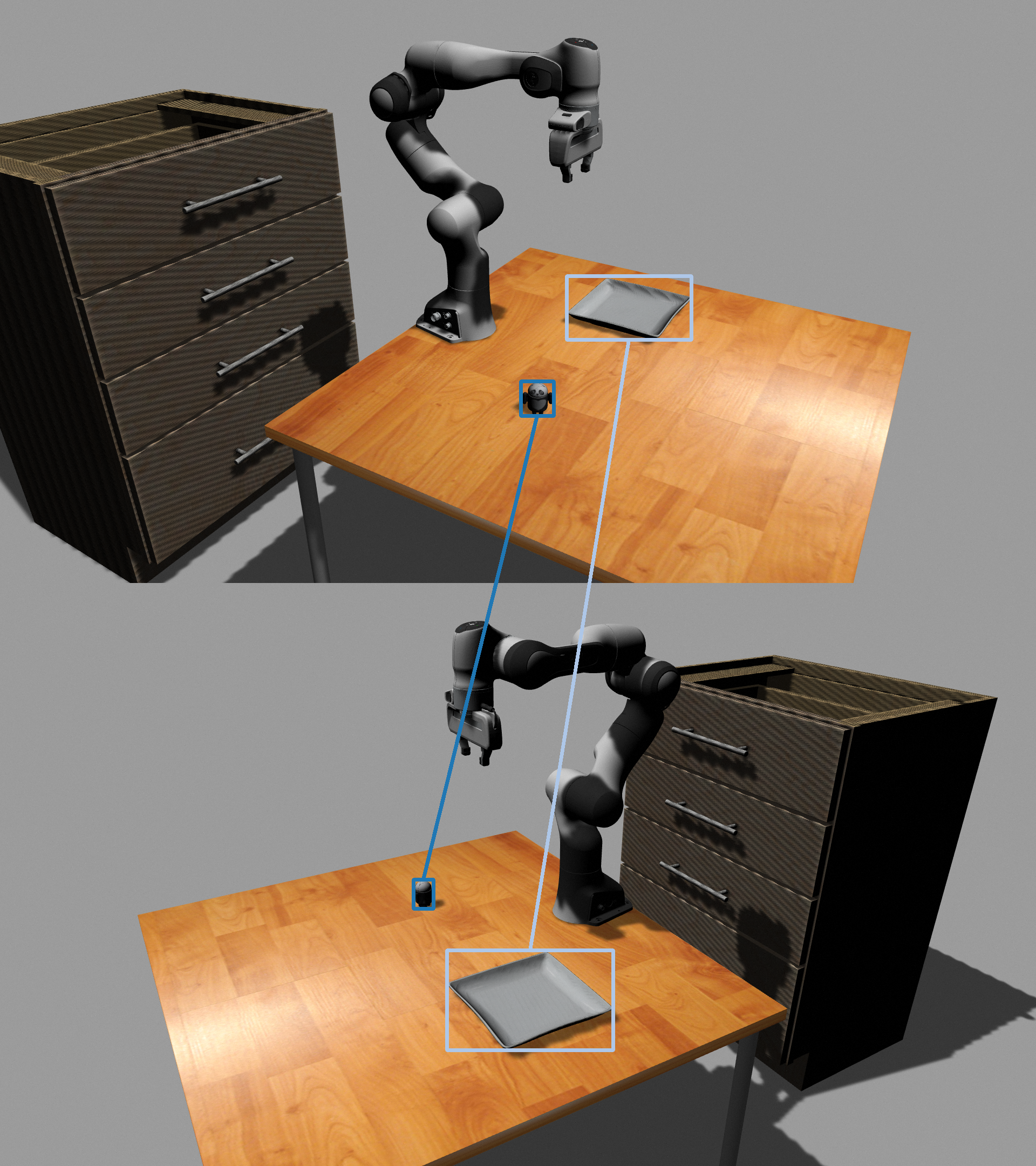}
   \caption{Matching results of 2D bounding boxes.}
   \label{fig:match}
\end{figure}
\begin{figure}[htbp]
  \centering
  % \fbox{\rule{0pt}{2in} \rule{0.9\linewidth}{0pt}}
\includegraphics[width=0.99\linewidth]{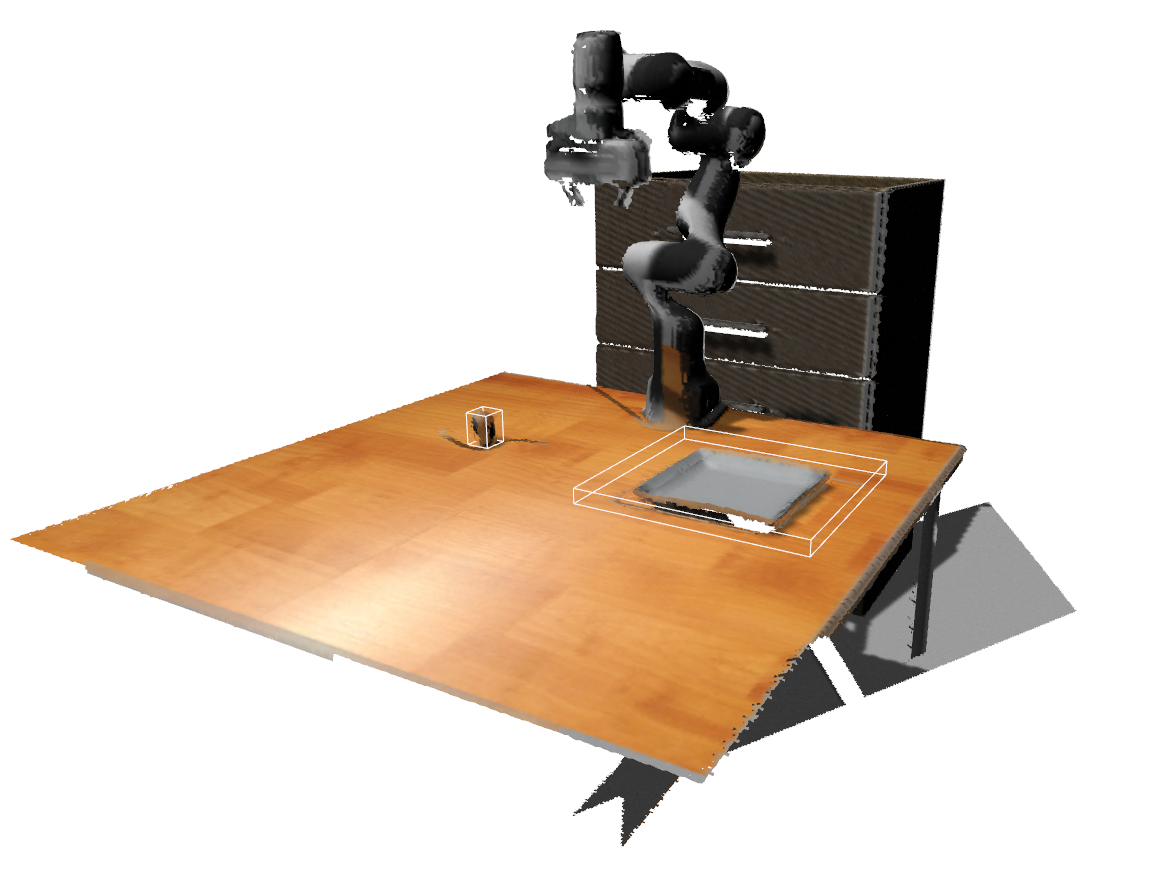}
   \caption{3D bounding boxes.}
   \label{fig:3dbbox}
\end{figure}
\begin{figure}[htbp]
  \centering
  % \fbox{\rule{0pt}{2in} \rule{0.9\linewidth}{0pt}}
\includegraphics[width=0.99\linewidth]{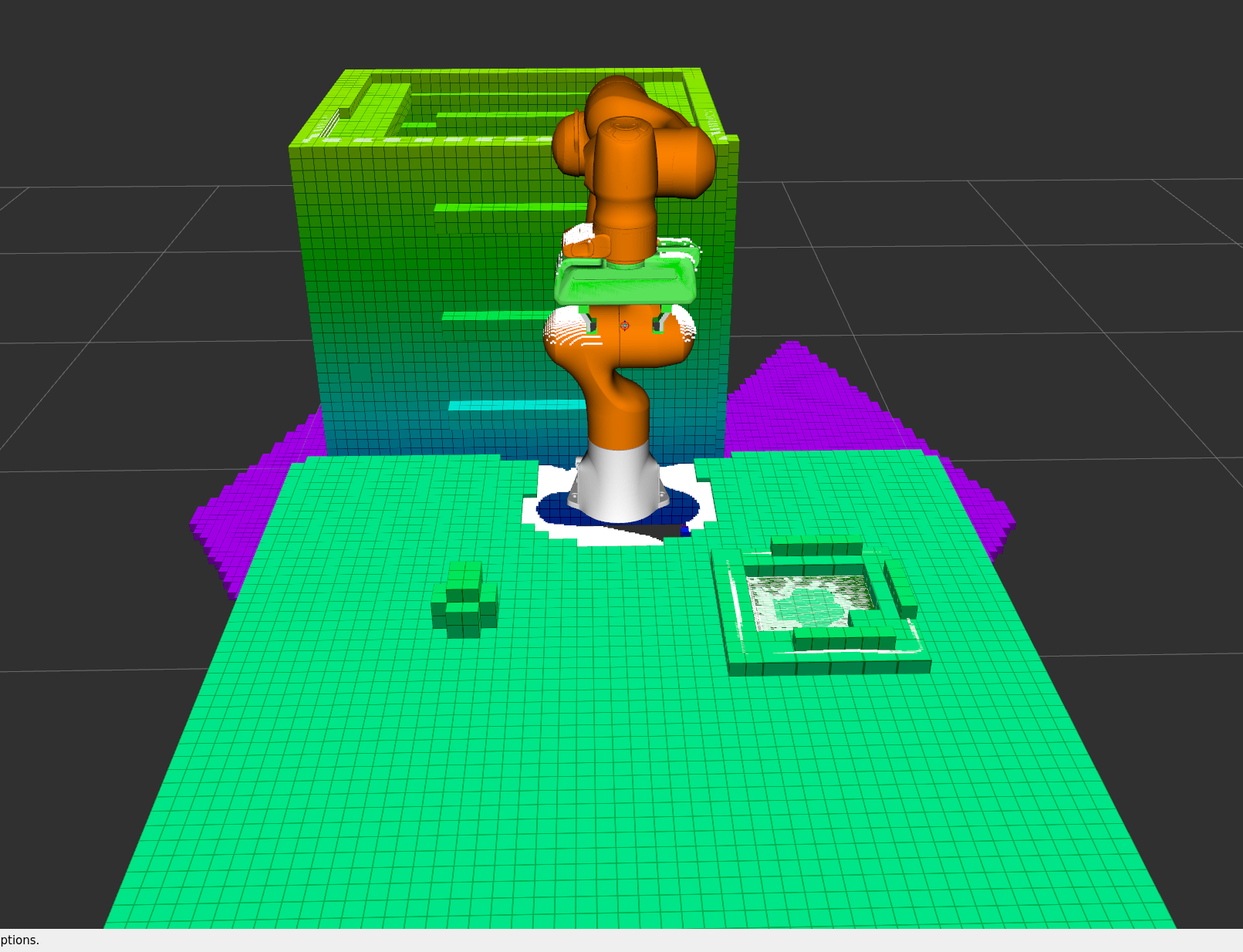}
   \caption{Visualization of the occupancy map. Occupancy maps represent the environment as a grid, where each cell can be free, occupied, or unknown.}
   \label{fig:occ_map}
\end{figure}
 The simulation framework is anchored in ROS\cite{koubaa2017robot} (Robot Operating System), utilizing Gazebo\cite{qian2014manipulation} for the realistic rendering of the environment and the physical interactions therein. The robotic arm's motion planning is facilitated by the integration of the MoveIt module, renowned for its comprehensive motion planning capabilities, and the OMPL\cite{sucan2012open} (Open Motion Planning Library), which offers a suite of advanced algorithms for efficient path planning and obstacle avoidance. The choice of the Franka arm's URDF (Unified Robot Description Format) model and the incorporation of the PartNet-Mobility Dataset's cabinet model necessitated a meticulous conversion and scaling process to ensure compatibility with the Gazebo environment. This setup's fidelity and functionality were rigorously validated through iterative testing phases, focusing on the robot's ability to physically  interact with the diverse objects. 
The agent processes three distinct RGBD images, utilizing the GLIP model for visual grounding, complemented by the reasoning outputs from large foundation models in our baseline. As shown in Figure \ref{fig:match}, multi-view 2D bounding boxes are aligned using geometric correspondences derived from depth data. These aligned bounding boxes are then fed into the 3D perception loop. It integrates multi-view RGBD images and 2D perception grounding results to generate 3D bounding boxes for objects as shown in Figure \ref{fig:3dbbox}, as well as their corresponding point clouds and occupancy maps as shown in Figure \ref{fig:occ_map}. Subsequently, the 3D point clouds are input into a grasp detection module, which is responsible for generating the grasp pose.
% \subsection{Planning with occupancy map}

\section{Implementation details of Vision Langauge Model}
\label{app_model}
We employ the same architecture as BLIP-2 \cite{li2023blip}. Specifically, we utilize the ViT-G/14 model from EVA-CLIP \cite{eva} and remove its last layer, using the output features of the second last layer instead. For the frozen language model, we adopt a pre-trained LLaMA-13B \cite{touvron2023llama} model and fine-tune the q-former and language model using the ShareGPT4-V dataset\cite{chen2023sharegpt4v}. We then utilize the obtained language model as the initial language model for multi-modal code-generation pre-training and special finetuning. Additionally, we convert the data type of parameters of the frozen ViT \cite{dosovitskiy2020image} and language model to FP16 during pre-training to increase efficiency.
\begin{table}[htbp]
\centering
\resizebox{0.99\linewidth}{!}{%
\begin{tabular}{l|l}
\hline
\textbf{Configuration Component} & \textbf{Details} \\
\hline
Model Type & RoboCodeX \\
\hline
\multicolumn{2}{l}{\textbf{Parameters}} \\
\quad - Hidden Size & 5120 \\
\quad - Attention Heads & 40 (Text), 16 (Vision) \\
\quad - Intermediate Size & 13824 (Text), 6144 (Vision) \\
\hline
Q-former & Standard Transformer \\
\quad - Dropout Prob & 0.1 \\
\quad - Encoder Hidden Size & 1408 \\
\quad - Hidden Size & 768 \\
\quad - Intermediate Size & 3072 \\
\quad - Attention Heads & 12 \\
\hline
Vision Model & ViT \\
\quad - Hidden Size & 1408 \\
\quad - Intermediate Size & 6144 \\
\quad - Attention Heads & 16 \\
\hline
Text Model & LLaMa \\
\quad - Hidden Size & 5120 \\
\quad - Intermediate Size & 13824 \\
\quad - Attention Heads & 40 \\
\hline
\end{tabular}
}
\caption{Model card of RobocodeX.}
\end{table}

The detailed implementation of the vision adapter, which takes the concatenated class tokens from different stage in vision transformer as input, is demonstrated in Listing \ref{lst:ros-adapter}.
\begin{lstlisting}[basicstyle=\tiny,caption={Pseudo-code of the Vision adapter}, label={lst:ros-adapter}]
class Adapter(nn.Module):
    def __init__(self, config):
        super().__init__()
        self.config = config
        self.activation_fn = ACT2FN["silu"]
        hidden_size = config.vision_config.hidden_size
        intermediate_size = hidden_size // 4
        output_size = config.qformer_config.hidden_size
        self.fc1 = nn.Linear(hidden_size, intermediate_size)
        self.fc2 = nn.Linear(intermediate_size, output_size)
        self.layernorm = nn.LayerNorm(output_size, eps=config.vision_config.layer_norm_eps)
    def forward(self, hidden_states: torch.Tensor) -> torch.Tensor:
        hidden_states = self.fc1(hidden_states)
        hidden_states = self.activation_fn(hidden_states)
        hidden_states = self.fc2(hidden_states)
        hidden_states = self.layernorm(hidden_states)
        return hidden_states
\end{lstlisting}

\section{Joint prediction of Articulated Objects}
\label{app:joints}
In this work, we adopt GAMMA \cite{yu2023gamma} as the base model and then fine-tune it using improved point clouds from multiple views. This is because GAMMA only uses the point cloud obtained from a single depth camera, resulting in domain gaps in our experimental setting. Except for the point cloud sources, all settings remain the same as in GAMMA. The GAMMA framework first analyzes articulated objects by segmenting them into distinct rigid parts and estimating their articulation parameters from a partial point cloud observation \(P=\{\mathbf{p}_{i}\in \mathbb{R}^{3}\}_{i=1}^{N}\). GAMMA utilizes a PointNet++ backbone for feature extraction and has three key components: segmentation, offset, and joint parameter estimation.

\begin{enumerate}
    \item \textbf{Segmentation Head:} It classifies each point into categories - static, revolute, or prismatic \(\left(\hat{c}_{i} \in \{0, 1, 2\}\right)\).
    \item \textbf{Offset Head:} It computes offset vectors \(\left(\hat{\mathbf{o}}_{i}\in \mathbb{R}^3\right)\), guiding points towards their part centroids.
    \item \textbf{Joint Parameter Head:} It predicts joint parameters, projecting each point onto the joint axis \(\left(\hat{\mathbf{v}}_{i}\in \mathbb{R}^3\right)\) and estimating the axis direction \(\left(\hat{\mathbf{d}}_{i}\in \mathbb{R}^3\right)\).
\end{enumerate}

The training involves minimizing a loss function, combining segmentation, offset, and joint parameter losses:

\[
\mathcal{L} = \frac{1}{N} \sum_{i}^N \Bigl[ \mathcal{L}_{c}(\hat{c}_{i},c_{i}) + \mathcal{L}_{o}(\hat{\mathbf{o}}_{i},\mathbf{o}_{i}) + \mathcal{L}_{v}(\hat{\mathbf{v}}_{i},\mathbf{v}_{i}) + \mathcal{L}_{d}(\hat{\mathbf{d}}_{i},\mathbf{d}_{i}) \Bigr],
\]

Here, \(\mathcal{L}_{o}(\hat{\mathbf{o}}_{i},\mathbf{o}_{i})\) is the offset loss, calculated as:

\[
\mathcal{L}_{o}(\hat{\mathbf{o}}_{i},\mathbf{o}_{i}) = \|\hat{\mathbf{o}}_{i}-\mathbf{o}_{i}\| - \Bigl(\frac{\mathbf{o}_{i}}{\|\mathbf{o}_{i}\|}_{2}\cdot \frac{\hat{\mathbf{o}}_{i}}{\|\hat{\mathbf{o}}_{i}\|}_{2}\Bigr)
\]

After training, we freeze parameters for feature inference. We perform clustering using DBSCAN, grouping points into part clusters based on the predicted semantics and axis-aware shifts \(\left(\mathbf{p}_{i} + \hat{\mathbf{o}}_{i}\right)\) and \(\left(\mathbf{p}_{i} + \hat{\mathbf{v}}_{i}\right)\). Each segmented part assists in voting for joint parameters.

The final performance on unseen testing  set is shown in Table \ref{tab:articulated}.
\begin{table}[h]
\centering
\caption{Evaluation on articulated objects joint prediction.}
\resizebox{0.99\linewidth}{!}{%
\begin{tabular}{c|c|c|c|c}
\midrule
\textbf{Objects} & \textbf{AP 75} & \textbf{Axis Direct Error} & \textbf{Axis Trans Error} & \textbf{Joint Type Acc (\%)} \\
\hline
Cabinet & 83.13 & 6.57 & 10.52 & 97.5 \\
% \hline
Microwave & 95.72 & 2.56 & 3.71 & 100 \\
% \hline
Refrigerator & 89.75 & 8.92 & 5.36 & 100 \\
\midrule
\end{tabular}
}
\label{tab:articulated}
\end{table}

\section{Grasp Pose Prediction}
\label{app:grasp}
\subsection{Grasp Pose Proposal Generation}
\label{sec:grasp-perception}

In the realm of robotic grasping, the effective generation of grasp poses is pivotal for successful object manipulation. This paper delineates two distinct models for Grasp Pose Proposal Generation, allowing an agent to choose based on preferences inferred by the advanced algorithmic framework, RoboCodeX.

The first model, Anygrasp, is documented in Fang et al. (2023)~\cite{fang2023anygrasp}. Anygrasp is a sophisticated pre-trained model adept at generating grasp poses. It utilizes a single RGB image and a corresponding point cloud to propose collision-free grasps, specifically tailored for a parallel jaw gripper. The model's efficacy lies in its ability to interpret complex scenes and propose viable grasping solutions.

The second model, herein referred to as the “central\_lift” method, takes a different approach by generating top-down grasp poses. This method simplifies the grasp generation process, focusing on the central lifting points of objects, which often align with their center of mass.

Further, to refine the grasp selection process, three specific preferences are incorporated into the function ``parse\_adaptive\_shape\_grasp\_pose()". It is crucial to select these preferences judiciously, as they vary significantly depending on the task requirements and the physical characteristics of the targeted objects. The preferences are as follows:
\begin{itemize}
\item \textbf{Preferred Position:} An Optional parameter, indicating the preferred position for the gripper tip point.
\item \textbf{Preferred Approach Direction:} An Optional parameter, denoting the preferred approach direction for the gripper.
\item \textbf{Preferred Plane Normal:} An Optional parameter, representing the preferred normal direction of the gripper's plane.
\end{itemize}

\subsection{Grasp Execution}
\label{sec:quality-control}

Upon the determination of an optimal grasp pose, the subsequent phase involves its execution. Leveraging the insights from Dasari et al. (2023)~\cite{dasari2023learning}, a simplified pre-grasp strategy is employed. Let $\overrightarrow{p}$ symbolize the identified grasp point, and $\overrightarrow{a}$ represent the approach vector as suggested by the grasping model. The trajectory followed by the robot gripper is mathematically expressed as:
\begin{equation*}
\langle \overrightarrow{p} - 0.1\overrightarrow{a},; \overrightarrow{p} - 0.08\overrightarrow{a},; \overrightarrow{p} \rangle
\end{equation*}

This trajectory indicates an approach towards the target object from a pre-defined position, progressively reducing motion increments to ensure precision and stability. The gradual deceleration is a critical aspect of the process, especially for handling lightweight or delicate objects that are susceptible to displacement or damage through abrupt movements.

The final phase of the grasp involves a closed-loop actuation of the gripper, ensuring a secure yet careful grip on the object. This process is meticulously controlled to avoid exerting excessive force, thereby preventing potential damage to the object being manipulated.

\section{Introduction of the APIs and the prompts}
\label{app:api and prompt}
The provided Python application programming interface (API) enables code generation for robot control and planning within robot operating systems. As demonstrated in Listing \ref{lst:ros-api} and Listing \ref{lst:ros-api1}, the APIs we provided contain modules for operating system interactions via ROS, robot perception utilities for sensing environments, and motion planning capabilities. Key perception functions determine object positions, bounding boxes, grasp poses, detected objects, and plane normals. Manipulation functions handle attaching, detaching, and moving objects. For motion, the API offers path planning and execution, including generating paths around joints and following predefined trajectories. Gripper poses and grasps can be controlled explicitly. As shown in Listing \ref{lst:ros-api2}, we also design prompts to guide large foundation models in adaptively generating robotic grasp poses. Our method leverages preferred positions, approach directions, and plane normals to create task-oriented gripper alignments and orientations. By using preferences rather than hard-coded values, our approach enables versatile grasping behavior tailored to specific tasks and objects. The rationale behind these preferences is crucial - they select appropriate grasps aligned and oriented to the task and target. Depending on complexity, additional helper functions may be required, definable in external scripts to ensure completion.
For safety and effectiveness, we establish boundaries within the robot's workspace. We also advocate avoiding hard-coded values in prompts where possible, instead using APIs and tools to dynamically adapt values to varying tasks and objects. This ensures responsiveness. 
Finally, we strictly emphasize that only Python code and comments starting with `\#' can be generated.

\begin{figure*}[htbp!]
  \centering
  % \fbox{\rule{0pt}{2in} \rule{0.9\linewidth}{0pt}}
\includegraphics[width=0.8\linewidth]{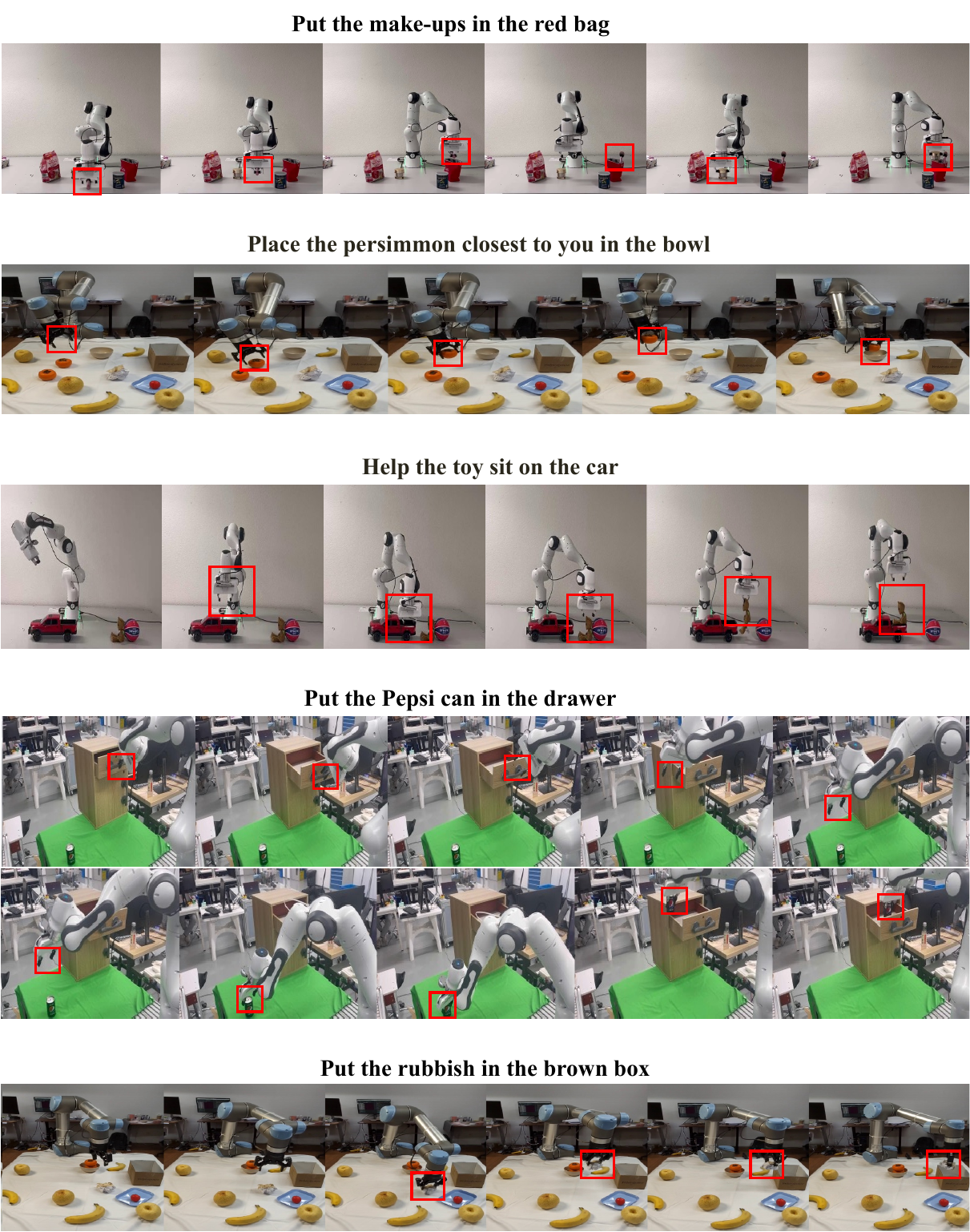}
   \caption{Real world experiments on Franka Emika Panda robot arm and UR5 robot arm.}
   \label{fig:real_app}
\end{figure*}

\section{Real world experiments}
To assess the cross-platform generalization capability of our RoboCodeX framework, we conducted experiments involving two distinct robot arms: the Franka Emika Panda robot arm and the UR5 robot arm, all without any task-specific fine-tuning tailored to each platform. Despite the inherent differences in environments and object sets between the two robotic systems, RoboCodeX delivered impressive results on both platforms by simply modifying the robot configuration file. As demonstrated in Figure \ref{fig:real_app}, the UR5 robot arm exhibited remarkable precision in tasks involving fruit and cosmetic manipulations, while the Franka Emika Panda robot arm adeptly positioned a toy on a car as per the given instructions.
The adaptability of our system across these robot arms, without the need for platform-specific training, underscores its versatility for multi-robot deployment. These zero-shot transfer experiments vividly showcase RoboCodeX's ability to generalize control policies effectively to new testbeds with minimal reconfiguration, making it a valuable asset for scalable real-world robot learning.

\begin{figure*}
\begin{lstlisting}[caption={Python API for Code Generation (part one)}, label={lst:ros-api}]
#You can use the following imported modules and functions in your script:

# Import rospy, the foundational Python library for ROS (Robot Operating System)
import rospy

# Import move_group, the class handle for moving the robot arm in MoveIt!
import move_group

# Import various ROS message types for handling robot pose and orientation
from geometry_msgs.msg import PoseStamped, Pose, Point, Quaternion

# Import utility functions for perception
from perception_utils import (
    get_object_center_position,  # Returns the position of an object in the world frame. Returns: position: np.array [x,y,z]
    get_object_pose,              # Returns the pose of an object in the world frame. Returns: pose: Pose
    get_3d_bbox,                 # Returns the 3D bounding box of an object in the world frame. Args: object_name: str. Returns: bbox: np.array [x_min, y_min, z_min, x_max, y_max, z_max]
    get_obj_name_list,           # Returns a list of names of objects present in the scene
    parse_adaptive_shape_grasp_pose, # Parse adaptive grasp pose for objects. Args: object_name: str, preferred_position: Optional(np.ndarray); preferred gripper tip point position; preferred_approach_direction: Optional(np.ndarray), preferred gripper approach direction; preferred_plane_normal: Optional(np.ndarray), preferred gripper plane normal direction. Returns: grasp_pose: Pose
    parse_central_lift_grasp_pose, # This method involves a vertical lifting action. The gripper closes at the center of the object and is not suitable for elongated objects and is not suitable for the objects with openings, as the gripper's width is really small. It is optimal for handling spherical and cuboid objects without any opening that are not ideal for off-center grasping. Args: object_name: str, description: Optional(str) in ['top', 'center'], Returns: grasp_pose: Pose
    parse_place_pose,            # Predict the place pose for an object relative to a receptacle. Args: object_name: str, receptacle_name: Optional(str), position: Optional(np.array) [x,y,z], . Returns: place_pose: Pose
    detect_objects,              # Detect and update task-relevant objects' status in the environment. Call this function before interaction with environment objects. Args: object_list: Optional(List[str]), objects to detect.
    get_object_joint_info,       # Get the joint info of an object closest to a given position. Args: obj_name: str, name of the object; position: np.ndarray, select the joint closest to this position; type: str, allowed type of the joint, choice in ['any', 'revolute', 'prismatic']. Returns: joint_info: dict, joint info of the object. {'joint_position':[x,y,z],'joint_axis':[rx,ry,rz],'type':str}
    get_plane_normal,            # Get the plane normal of an object closest to the given position. Args: obj_name: name of the object position: np.ndarray, select the plane closest to this position. Returns: np.ndarray, normal vector [x,y,z]
)
\end{lstlisting}
\end{figure*}

\begin{figure*}
\begin{lstlisting}[caption={Python API for Code Generation (part two)}, label={lst:ros-api1}]
# You can use the following imported modules and functions in your script:

# Import utility functions for robot motion planning and execution
from motion_utils import (
    attach_object,  # Attaches an object to the robot gripper in the planning space. Call this function right after closing the gripper. Args: object_id: str. 
    detach_object,   # Detaches an object from the robot gripper in the planning space. Call this function right after opening the gripper. Args: object_id: str. 
    open_gripper,    # Open the gripper. No args.
    close_gripper,   # Close the gripper. No args.
    move_to_pose,    # Move the gripper to pose. Args: pose: Pose
    move_in_direction, # Move the gripper in the given direction in a straight line by certain distance. Note that you can use the surface normal and the joint axis. Args: axis: np.array, the move direction vector ; distance: float.
    generate_arc_path_around_joint, # Generate a rotational gripper path of poses around the revolute joint. Args: current_pose: Pose, current pose of the gripper; joint_axis: np.array, the joint axis of the revolute joint; joint_position: np.array, the joint position of the revolute joint; n: int, number of waypoints; angle: float, angle of rotation in degree. Returns: path: List[Pose]
    follow_path,     # Move the gripper to follow a path of poses. Args: path: List[Pose]
    get_gripper_pose, # Get the gripper pose. No args. Returns: pose: Pose
    grasp,           # Executes a grasp motion at the grasp_pose. Args: grasp_pose: Pose
)
You are encouraged to use above APIs to complete the task.
\end{lstlisting}
\end{figure*}

\begin{figure*}
\begin{lstlisting}[caption={Other Prompts for Code Generation}, label={lst:ros-api2}]
# There are three preferences for function parse_adaptive_shape_grasp_pose().
# Note that you need to choose the right preferences for different tasks and objects.
# preferred_position: Optional(np.ndarray), preferred gripper tip point position. 
# preferred_approach_direction: Optional(np.ndarray), preferred gripper approach direction.
# preferred_plane_normal: Optional(np.ndarray), preferred gripper plane normal direction.
# This preference selects the grasp pose so that the gripper is parallel to the plane defined by the normal.
# It also means the gripper grasping at an axis with perpendicular pose to the axis. 
# Note that you may always need to create your arbitrary functions to help you complete the task, which will be defined by external scripts.
# The robot working space on table top is in range [-0.5, 0.2] in x-axis and [-0.5, 0.5] in y-axis. The height of the table top is 1.05.
# Note that you are a strict coder, you can not hard code a predefined value, but you need to use api and tools to detect the value.
# Please pay attention to the description and specific requirements of the target object in the task. You may need to write some founctions by yourself to determine whether the requirements are met.
# Your generated content should only contain comments starting with '#' and python code!

\end{lstlisting}
\end{figure*}

\end{document}